%% file: p3_capacity.tex
\documentclass{article}
\pdfoutput=1
\usepackage[preprint]{neurips_2024}
\usepackage[utf8]{inputenc}
\usepackage[T1]{fontenc}
\usepackage[hidelinks]{hyperref}
\usepackage{url}
\usepackage{booktabs}
\usepackage{amsfonts}
\usepackage{amsmath}
\usepackage{amssymb}
\usepackage{nicefrac}
\usepackage{microtype}
\usepackage{graphicx}
\usepackage{xcolor}
\usepackage{enumitem}
\usepackage{tikz}
\usetikzlibrary{arrows.meta}
\graphicspath{{figs/}{figs_p456/}}
\newcommand{\kstar}{k^{\ast}}
\newcommand{\ddecl}{d_{\mathrm{decl}}}
\input{numbers}
\input{numbers_addenda}
\definecolor{housetea}{HTML}{1F6F78}
\definecolor{houseclay}{HTML}{C0442E}

\title{In-Context Binding Capacity in Language Models}

\author{%
  Manas Venkata Sai Ravulapalli\\
  Efficient Computation Inc.\\
  \texttt{manas@perseus.so}
  \And
  Samrath Chadha\\
  Efficient Computation Inc.\\
  \texttt{samrath@perseus.so}
}

\begin{document}
\maketitle

\begin{abstract}
How many assignments can a language model recall before it loses track of which value belongs to which entity? We measure this limit using continuous recall curves for 12 models at or below 3B parameters and a threshold sweep over \NModelsRaw{} open models up to 12B. On the continuous curves, the load at which recall falls halfway to chance follows $K_{50}=cN^{\alpha}$, with $\alpha=\CapAlpha$ and $R^2=\CapRsq$. The broader sweep shows an $\KstarRangeFactor\times$ range associated with pretraining recipe, although the continuous curves show no detectable recipe effect after controlling for scale, with few modern models in the fit. We derive why interference can lower measured capacity by reducing single-binding recall even when the load-dependent recall profile is unchanged. Direct task training also exceeds the extrapolated zero-shot law, but different measurement criteria prevent interpreting that comparison as a capacity gain. Its formation times follow a power-law form in two independent codebases, conditional on runs that succeed. Together, these results characterize capacity at the model's query interface. Bounds on joint recall and a decomposition of policy errors connect this measurement to working memory and instruction following, without treating recall as a measure of alignment. The controlled task also provides a baseline for testing whether binding limits constrain world-state tracking; the present experiments do not measure state updates or downstream transfer.
\end{abstract}

\section{Introduction}
A language model that reads a configuration file must remember which value was assigned to which
variable, and remember it for every variable at once. This is an example of in-context \emph{binding}. Like human working memory, it has a capacity: past some number of simultaneous bindings, recall collapses.

The same problem arises when an assistant tracks a user's constraints across turns, a program's
variables across execution, or objects as actions change their state. These tasks also require
updating facts and using the current state to answer or act. Our task isolates the static retrieval
step: it asks whether the model returns the value assigned to the queried entity. Measuring that
step can make later tests of state tracking more precise, but it does not establish that a model
maintains a world model.

Parameter count is a natural capacity predictor. On continuous recall-versus-$K$ curves at $\le$3B, capacity scales with parameters as $K_{50} = c\,N^{\alpha}$, with $\alpha=\CapAlpha$. Yet on the instrument that spans the largest models, the Pythia ladder holds $\kstar\in[\PythiaKstarLo,\PythiaKstarHi]$ from 410M to 12B, a $29\times$ increase in parameters, while the 3B Qwen2.5 reaches the measurement ceiling. This contrast is associated with recipe. Both instruments measure capacity, but over different model ranges and at different resolution (Section~\ref{sec:setup}).

Binding capacity separates into what pretraining delivers for zero-shot use, how that use withstands interference, and what direct task training can build. Our capacity fit describes a zero-shot operating regime, whereas the scaling laws of \citet{kaplan2020} and \citet{hoffmann2022} describe language-model loss. Direct task training provides a separate measure of learnability. Section~\ref{sec:boundary} reports a factor of $\BoundFactor$ relative to an extrapolation of the zero-shot law, with different estimands and a prediction outside the task's load domain. The independent re-measurement gives a post-hoc zero-shot scaling law on models at or below 3B, and the broad sweep shows recipe-associated differences across model ladders. The geometric comparisons do not establish a mechanism: recipe contrasts and trial-level error prediction answer different questions (Section~\ref{sec:geometry-evidence}).

The contribution is a measured capacity law together with conditions for interpreting it.
The threshold analysis shows when an exponent transfers to another recall requirement and when
model order can change (Section~\ref{sec:thresholds}). The interference identity separates changes
in single-binding reliability from changes in the remaining load profile
(Section~\ref{sec:normalisation}). These results motivate comparisons at a stated absolute recall
requirement (Section~\ref{sec:use}), while the geometric and formation analyses delimit the
mechanistic and training claims that the measurements support. Section~\ref{sec:implications}
relates these findings to working-memory tasks, sets out a test of world-state tracking, and
derives the additional assumptions needed to connect recall reliability to policy adherence.

\section{Measuring capacity}\label{sec:setup}
\paragraph{Task.} Each trial presents $K$ bindings ``\texttt{$e_i$: $o_i$}'' (entity $e_i$, single-token
obligation $o_i$, from disjoint pools), then an interference block of $D$ distractor tokens, then the
query ``\texttt{The task for $e_j$ is:}'' for a random $j$. The model is correct iff its highest-logit obligation token is $o_j$. Varying $K$ probes \emph{capacity}; varying $D$ probes \emph{interference-robustness}. Chance
over the obligation pool is $1/|\mathcal{O}|=\Chance$. When the $K$ obligations are distinct, choosing
uniformly among those present answers the query correctly with probability $1/K$. Because $K$ varies along the capacity axis, the capacity
statistics below are computed against $1/|\mathcal{O}|$.

\begin{figure}[t]
\centering
\begin{tikzpicture}[x=1cm, y=1cm, line width=0.5pt, font=\scriptsize]
\foreach \i/\xa in {1/0, 2/1.4} {
  \draw[draw=housetea, fill=housetea!10] (\xa, 1.1) rectangle (\xa+1.3, 1.75);
  \node at (\xa+0.65, 1.425) {$e_{\i}{:}\,o_{\i}$};
}
\node at (3.0, 1.425) {$\cdots$};
\draw[draw=housetea, fill=housetea!10] (3.3, 1.1) rectangle (4.6, 1.75);
\node at (3.95, 1.425) {$e_K{:}\,o_K$};
\draw[draw=black!45, fill=black!7] (4.8, 1.1) rectangle (7.4, 1.75);
\node at (6.1, 1.425) {$D$ distractor tokens};
\draw[draw=houseclay, fill=houseclay!10] (7.6, 1.1) rectangle (9.7, 1.75);
\node at (8.65, 1.425) {``task for $e_j$ is:''};
\draw[housetea, |-|] (0, 0.72) -- (4.6, 0.72);
\node[housetea, anchor=north] at (2.3, 0.62) {capacity axis: vary $K$};
\draw[black!60, |-|] (4.8, 0.72) -- (7.4, 0.72);
\node[black!60, anchor=north] at (6.1, 0.62) {robustness axis: vary $D$};
\node[houseclay, anchor=north, align=center] at (8.65, 0.62) {correct iff the top\\ logit is $o_j$};
\begin{scope}[shift={(10.6, 0)}]
\draw[->] (0, 0) -- (3.35, 0) node[anchor=north east, inner sep=2pt] {$K$};
\draw[->] (0, 0) -- (0, 1.95);
\node[rotate=90, anchor=south] at (-0.15, 0.95) {recall};
\draw[housetea, line width=0.8pt] plot[smooth] coordinates
  {(0.15, 1.7) (0.8, 1.65) (1.3, 1.5) (1.8, 1.05) (2.3, 0.55) (2.9, 0.35)};
\draw[black!50, dashed] (0, 1.02) -- (3.2, 1.02);
\node[black!50, anchor=east, inner sep=1pt] at (3.2, 1.22) {midpoint};
\draw[black!50, dashed] (1.82, 0) -- (1.82, 1.05);
\node[anchor=north] at (1.82, -0.02) {$\kstar,\,K_{50}$};
\end{scope}
\end{tikzpicture}
\caption{\textbf{The task and its two axes.} Each trial presents $K$ entity--obligation bindings, a
$D$-token interference block, and a recall query; the model is correct iff its highest-logit obligation
token matches the queried binding. Varying $K$ probes capacity, varying $D$ probes
interference-robustness. Right: both capacity statistics are midpoint crossings of the recall curve
between the model's own $K{=}1$ ceiling and chance: $\kstar$ on the measurement grid and $K_{50}$ under a
logistic fit to the continuous curves (Section~\ref{sec:setup}).}
\label{fig:task}
\end{figure}
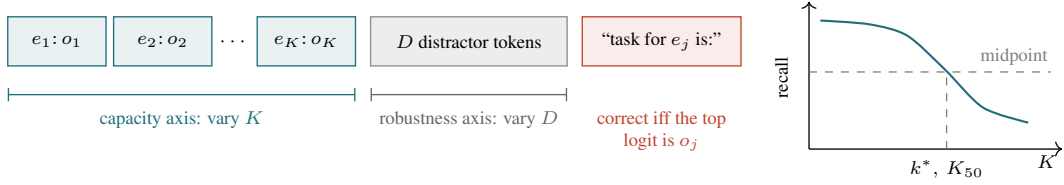

\paragraph{The capacity statistics, and their pitfall.} $\kstar$ is the smallest $K$ at which recall falls below the midpoint between the $K=1$ ceiling and chance. It is a threshold crossing of a smooth curve, and thresholding can manufacture discontinuities where the underlying quantity varies continuously \citep{mirage2304}.
We therefore treat $\kstar$ as a \emph{summary statistic} and not as evidence of a sharp transition. The released table contains $\kstar$ and the geometry but not the full recall-versus-$K$ curves, so we cannot rule out the thresholding artifact from it directly.
Instead, we exclude any model whose $K=1$ ceiling is near chance (its threshold is noise), report the median difference rather than the range, and cluster on families.
Section~\ref{sec:law} releases the curves for twelve of the models. $K_{50}$ is the continuous analogue of $\kstar$ on those released curves: the load at which retained performance $r(K)$ (recall normalised between the model's own $K{=}1$ ceiling and chance) crosses one half under a logistic fit in $\log K$. 
We report right-censoring rather than clipping it, and validate the estimator on synthetic curves with known answers before touching data.

\paragraph{Broad-sweep exclusions.} \emph{(i)} $\kstar$ is defined as a threshold crossing
relative to a model's own $K=1$ ceiling. If that ceiling is itself at chance, the threshold is noise. One
model, Pythia-70M, has a $K=1$ accuracy of $\PythiaSeventyCeiling$ against a chance of $\Chance$: it cannot hold a
\emph{single} binding. This model is excluded, leaving $n=\NModels$, as it is the source
of the $12\times$ range in the raw table; without it, the range is $\KstarMin$ to $\KstarMax$, i.e.\ $\KstarRangeFactor\times$.

\emph{(ii)} Three models sit at $\kstar=\KstarMax$, the ceiling imposed by the entity pool, and are \emph{right-censored}: OLMo-1-7B, Qwen2.5-3B, and, informatively, the old-recipe OPT-6.7B. The low end of the range is measured.

\paragraph{Models.} The released capacity table covers \NModelsRaw{} open models from ten families. \emph{Old} (pre-2023)
\emph{recipe} ($n=\NOldRaw$): GPT-2 (M/L/XL) \citep{radford2019}, GPT-Neo (125M, 1.3B), OPT (350M, 1.3B, 2.7B, 6.7B), BLOOM (560M, 1.7B), Cerebras-1.3B,
and the Pythia ladder (70M--12B) \citep{biderman2023}. \emph{Modern} ($n=\NModernRaw$): OLMo-1-7B, OLMo-2-7B with its SFT/DPO/Instruct ladder,
OLMoE, Qwen2.5 (0.5B--7B), Mistral-7B, DeepSeek-Coder-6.7B. All measurements are on frozen models. The
continuous re-measurement of Section~\ref{sec:law} covers twelve models at or below 3B: seven from
this table (OPT-350M/1.3B, Pythia-410M/1.4B/2.8B, Qwen2.5-0.5B/1.5B) together with OPT-125M,
Pythia-160M, Pythia-1B, SmolLM2-360M and OLMo-2-1B. The uncensored fit spans four families. The interference sweep
of Section~\ref{sec:robust} covers the eight models with committed $D$-sweep curves.

\section{A zero-shot scaling law for capacity}\label{sec:law}
\subsection{The continuous instrument}
A re-measurement with committed code re-runs the models on one instrument and releases the full recall-versus-$K$ curves.
A first wave found the GPT-2 family and Pythia-70M below the $K{=}1$ exclusion floor; a second ran the twelve models above
that floor on a $K$ grid to 64 with six seeds per cell. The GPT-2 result disagrees with the original instrument, where those
models hold a measurable $\kstar$; we report it as an instrument difference. Models censored at the first wave's $K{=}24$
grid resolve to finite values on the full grid (OPT-1.3B to $\TrnOptResolved$).

\subsection{The \texorpdfstring{$K_{50}$}{K50} power law}
Over these curves, capacity
fits a power law, $K_{50} = c\,N^{\alpha}$ with $\alpha=\CapAlpha$ ($R^2=\CapRsq$), family-clustered CI
$[\CapAlphaCILo, \CapAlphaCIHi]$, leave-one-family-out range $[\CapLofoLo, \CapLofoHi]$
(Figure~\ref{fig:capacitylaw}). 
The interval contains proportionality and excludes zero, and removing any single model moves $\alpha$ by at most $\CapLooMax$. The fit uses the
$\CapFitN$ of twelve models with an uncensored $K_{50}$ (OLMo-2-1B is right-censored at the grid
maximum). The uncensored models span four families, so the clustered interval and the leave-one-family-out
range rest on four clusters.

\subsection{Recipe after controlling for scale}
At $\le$3B parameters, the recipe split of Section~\ref{sec:capacity} does not survive controlling for scale. Regressing $\log K_{50}$ on recipe and $\log$ parameters over the re-measured curves gives a recipe coefficient of $\TrnRecipeCoef$ (family-clustered CI
$[\TrnRecipeCILo, \TrnRecipeCIHi]$, spanning zero) and a scale coefficient of $\TrnScaleCoef$ (CI $[\TrnScaleCILo, \TrnScaleCIHi]$, excluding it).
The regression uses $n=\CapFitN$ models across four families, of which three are modern
(Qwen2.5-0.5B, Qwen2.5-1.5B, SmolLM2-360M); the null is correspondingly weak.
The threshold sweep reaches 12B, whereas this continuous re-measurement stops at 3B; continuous curves above that would show whether the recipe span reappears.

\begin{figure}[t]
\centering
\includegraphics[width=0.8\linewidth]{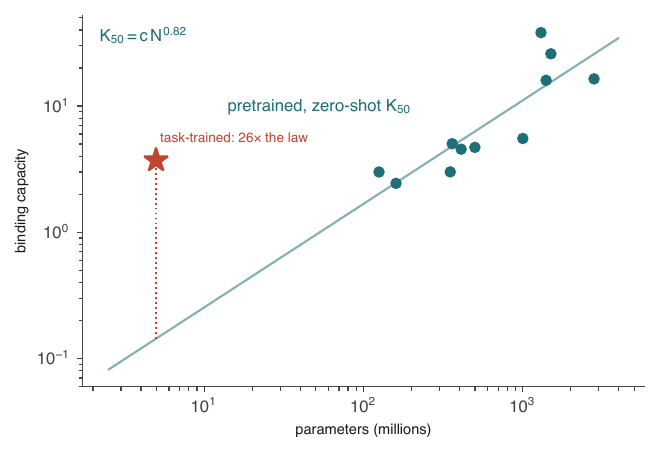}
\caption{\textbf{The $K_{50}$ scaling law and a comparison outside its fitted range.}
Circles: $K_{50}$ for the $\CapFitN$ uncensored re-measured models; the line is the fitted
$K_{50} = c\,N^{\CapAlpha}$. Star: a 4.98M-parameter transformer trained directly on the task, at
$\BoundFactor\times$ the law's prediction for its size (dotted riser). The two estimands differ (largest
learnable $K$ versus zero-shot $K_{50}$). The extrapolated prediction falls below one binding, so
the ratio is an extrapolation discrepancy rather than a measured capacity gain.}
\label{fig:capacitylaw}
\end{figure}

The line in Figure~\ref{fig:capacitylaw} describes how the midpoint of a zero-shot recall
curve varies with parameter count in this cohort. It does not imply better recall at every
load: two models can differ in their single-binding ceiling or in how sharply recall falls.
Those differences matter for a fixed absolute-recall requirement
(Section~\ref{sec:use}), while slope differences can change a crossing at a stricter
relative criterion (Section~\ref{sec:thresholds}). The broader threshold sweep includes
larger models, but its load grid and right-censored values cannot resolve a continuous
size law in that range. Its recipe contrast and the fitted line therefore describe
different comparisons, not two estimates of one effect.

\subsection{Sensitivity to the summary statistic}\label{sec:sensitivity}
The following checks reuse the existing corpus and are exploratory: none was predicted before
inspection of the main results. They test whether the scale association depends on one summary,
without treating a different summary as a replication of the same exponent.

\paragraph{The scale association persists in a threshold-free summary.} $K_{50}$ is
a threshold crossing, and one of twelve models is right-censored and excluded from the fit. Both
choices can be removed at once. The area under the retained-performance curve (AUC) needs no threshold and
is defined for every model, censored or not. Regressing $\log$ AUC on $\log$ parameters gives an
exponent of $\AucExp$ with $R^2=\AucRsq$ over all \AucN{} models, family-clustered CI $[\AucCILo,
\AucCIHi]$; restricting to the \CapFitN{} uncensored models gives $\AucExpUnc$ with $R^2=\AucRsqUnc$.
Both fits have a higher in-sample $R^2$ than the $K_{50}$ law's $R^2=\CapRsq$, although the dependent
variables differ, so this does not rank their predictive validity. The exponents are not comparable
in magnitude to $\alpha=\CapAlpha$: AUC integrates performance over a finite load range, whereas
$K_{50}$ locates a crossing. What carries over is the positive association with scale and its
interval excluding zero, with no threshold applied and no model discarded. AUC can nevertheless
saturate when the curve moves beyond the measured range; it does not identify unobserved capacity.

\paragraph{Single-binding recall is usually high.} A model with a low $K{=}1$ ceiling
could show a low $K_{50}$ for reasons unrelated to capacity. \CeilNAbove{} of \CeilN{} models hold a
ceiling between $\CeilLo$ and $\CeilHi$; the exception is Pythia-160M at $\CeilOutlier$. Ceiling is
only weakly related to scale ($R^2=\CeilRsq$), but this summary alone does not identify how ceiling uncertainty affects the fitted midpoints.

\paragraph{The fitted collapse is gradual, and its steepness covaries with capacity.} Fitting a logistic in
$\log K$ gives a slope per model, and those slopes span $\SlopeLo$ to $\SlopeHi$ over \SlopeN{}
models. A finite-slope logistic represents a smooth transition by construction; fitting it does not
independently exclude a sharper underlying change. The released curves permit that question to be
examined beyond the thresholded table \citep{mirage2304}. We looked
for a second scaling law in the slope and did not find one: slope is strongly coupled to $K_{50}$
($r=\SlopeCorrK$ in logs), and once $K_{50}$ is known, parameters add almost nothing (partial
$r=\SlopePartialN$). We report the range and not a law.

\subsection{When a scaling exponent transfers across thresholds}\label{sec:thresholds}
A scaling law at the midpoint leaves open whether size helps equally at a stricter recall requirement.
For positive load $K$, midpoint $K_{50,i}$, and parameter count $N_i$, let the fitted
retained-performance curve for model $i$ be
\begin{equation}
 \widehat r_i(K)=\frac{1}{1+\exp\{\beta_i(\log K-\log K_{50,i})\}},
 \qquad \beta_i>0.
 \label{eq:logistic-profile}
\end{equation}
For a retained-performance criterion $q\in(0,1)$, solving $\widehat r_i(K_{q,i})=q$ gives
\begin{equation}
 \log K_{q,i}=\log K_{50,i}+\frac{1}{\beta_i}\log\frac{1-q}{q}.
 \label{eq:threshold-transfer}
\end{equation}
Thus a common slope $\beta_i=\beta$ and an exact midpoint law $K_{50,i}=cN_i^{\alpha}$ imply
$K_{q,i}=c[(1-q)/q]^{1/\beta}N_i^{\alpha}$: changing the criterion changes the coefficient but
preserves the exponent. This is a sufficient condition, not an assumption established by our fits.

More generally, let $x_i=\log N_i$ and fit ordinary least squares with an intercept over the
\emph{same} uncensored models. Taking covariance and variance over that model set, linearity of
the regression slope in its response gives
\begin{equation}
 \widehat\alpha_q=\widehat\alpha_{50}
 +\log\frac{1-q}{q}\,
 \frac{\operatorname{Cov}(x_i,\beta_i^{-1})}{\operatorname{Var}(x_i)},
 \qquad \operatorname{Var}(x_i)>0.
 \label{eq:threshold-exponent}
\end{equation}
This follows by substituting Equation~\eqref{eq:threshold-transfer} into
$\widehat\alpha_q=\operatorname{Cov}(x_i,\log K_{q,i})/\operatorname{Var}(x_i)$.
The midpoint exponent therefore need not describe high-reliability recall. Slope variation can
change the fitted exponent when inverse slope covaries with size. The correction vanishes at
$q=1/2$, and also vanishes at every criterion if that covariance is zero. Changing the model set
because a different threshold becomes censored adds a separate selection effect.

These identities apply to the fitted logistic, which does not pass through
$\widehat r_i(1)=1$. They must not be substituted for the exactly anchored normalisation in
the empirical definition of $r(K)$ in Section~\ref{sec:setup}. The existing slope--capacity association motivates
checking transfer across criteria; it does not determine the covariance in
Equation~\eqref{eq:threshold-exponent} or provide a measured correction to $\alpha$.

\paragraph{Model order can depend on the required recall.}
For models $i$ and $j$, let $\delta_{ij}=\log K_{50,i}-\log K_{50,j}$ and
$h_{ij}=\beta_i^{-1}-\beta_j^{-1}$. Subtracting Equation~\eqref{eq:threshold-transfer} gives
\begin{equation}
 \log\frac{K_{q,i}}{K_{q,j}}
 =\delta_{ij}+h_{ij}\log\frac{1-q}{q}.
 \label{eq:ranking-threshold}
\end{equation}
When $h_{ij}\ne0$, the formal order changes at
$q^*=[1+\exp(-\delta_{ij}/h_{ij})]^{-1}$. Equal slopes preserve the order at every threshold;
equal slopes and midpoints give identical fitted curves. With unequal slopes, a reversal is relevant
to the measurements only if its crossing load lies inside both measured domains. This is a
conditional consequence of the fitted family, not an observed reversal in our models. It shows why
ranking models by their midpoint need not rank them by high-reliability retained performance.

\section{The threshold sweep to 12B}\label{sec:capacity}
\begin{figure}[t]
  \centering
  \includegraphics[width=\linewidth]{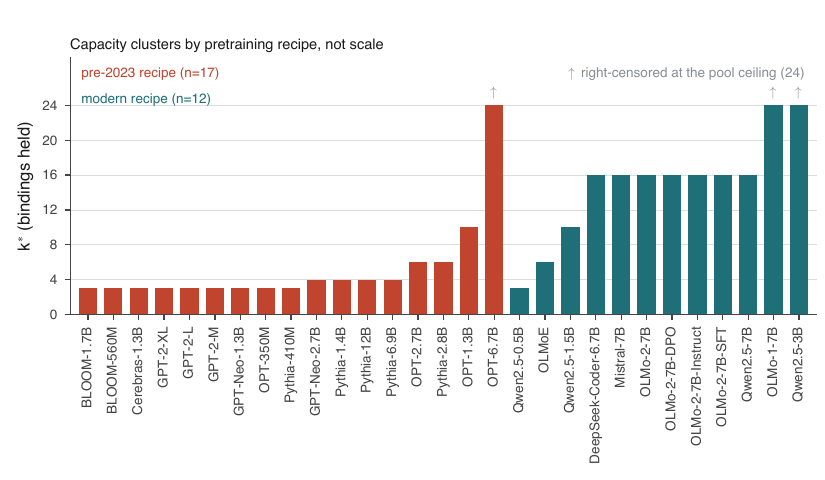}
  \caption{\textbf{Binding capacity $\kstar$ clusters by pretraining recipe.} Median $\kstar$ is $\KstarOldMed$ for old
  recipes and $\KstarModMed$ for modern ones (family-clustered $95\%$ CI of the difference $\KstarFamCI$; family-level
  $p=\KstarFamP$). The classes \emph{overlap}: OLMoE (modern) sits at $\KstarOlmoE$ and OPT-6.7B (old) reaches the censored
  ceiling. Arrows mark models right-censored at the entity-pool ceiling ($\kstar=\KstarMax$). Pythia-70M is excluded
  due to its $K=1$ accuracy being at chance.
  }
  \label{fig:capacity}
\end{figure}

Figure~\ref{fig:capacity} shows $\kstar$ across the \NModels{} measurable models. It spans $\KstarMin$ to
$\KstarMax$. Old recipes have median $\KstarOldMed$ ($n=\NOldModels$), compared with $\KstarModMed$ for
modern recipes ($n=\NModernModels$). Clustering the \NModelsRaw{} models into ten families, the median difference has 95\% CI $\KstarFamCI$, and
the family-level test gives $p=\KstarFamP$. The model-level Mann--Whitney test, which treats siblings as
independent draws, gives $p=\KstarModelP$.

\subsection{Recipe and scale}
 The Pythia ladder holds $\kstar=\PythiaLadder$ at 410M,
1.4B, 2.8B, 6.9B and 12B: flat across a $29\times$ range of parameters. Within a \emph{modern} recipe, scale
does help: Qwen2.5 goes $\QwenLadder$ at 0.5B, 1.5B, 3B and 7B. These ladders show a
recipe-associated contrast between scale trends, rather than an identified causal interaction. The Qwen ladder is \emph{not monotone}: the 3B
model reaches the censored ceiling and the 7B model sits below it.
Alignment leaves this threshold unchanged: the OLMo-2 base/SFT/DPO/Instruct ladder holds $\kstar=\KstarOlmoLadder$ at every stage. \citet{kim2405} found no consistent alignment benefit for entity tracking; this ladder shows no change detectable by the capacity threshold. Identical grid crossings do not establish identical recall curves.

\subsection{Overlap between recipes}
 Modern recipes do not uniformly
dominate. OLMoE, a modern model, sits at $\kstar=\KstarOlmoE$; OPT-1.3B, an old one, reaches $\KstarOptOneThree$; and OPT-6.7B, also
old, reaches the censored ceiling at $\KstarMax$. Both classes span the full $\KstarMin$--$\KstarMax$ range. The recipe effect is a
difference in medians ($\KstarOldMed$ versus $\KstarModMed$) that survives family clustering.

\subsection{Sensitivity to the disputed family}\label{sec:recipe-sensitivity}
This exclusion check is exploratory and uses the existing threshold table.
\paragraph{The recipe split does not rest on the disputed family.} The continuous instrument places
the GPT-2 family below the $K{=}1$ exclusion floor (Section~\ref{sec:law}), while the threshold sweep
assigns all \GptTwoDropped{} GPT-2 models $\kstar=\KstarOldMed$, the old-recipe median. Dropping them
moves the old-recipe median from $\KstarOldMed$ to $\GptTwoOldMed$ and the median difference from
$+13$ to $+\GptTwoDiff$. The split survives the instrument disagreement.

\section{The boundary: direct task training}\label{sec:boundary}
\subsection{Direct task training}
The task-trained comparison tests the scope of the power law. A 4.98M-parameter transformer trained directly on the
task holds an interpolated $\BoundMeasured$ bindings where the law predicts $\BoundPredicted$, a
factor of $\BoundFactor$, read over 400 from-scratch runs in that cell from a clean corpus of 423.
The estimands differ: one interpolates the fraction of training runs that reach a formation criterion,
while the other measures a midpoint of zero-shot recall. The denominator is also extrapolated below
the measured model-size range and below one binding. Since normalisation sets $r(1)=1$, a continuous
decreasing recall curve can reach its midpoint only at $K>1$. Thus a power law with $c>0$ and
$\alpha>0$ cannot extend to arbitrarily small models: $cN^{\alpha}<1$ leaves the domain of the
capacity definition. The factor above is an extrapolation discrepancy, not a measured training gain
or a bound on unused architectural capacity. The task-trained result establishes learnability in its
own setting; it does not calibrate the zero-shot law outside its observed range.

\subsection{The formation-time law}
\begin{figure}[t]
\centering
\includegraphics[width=0.7\linewidth]{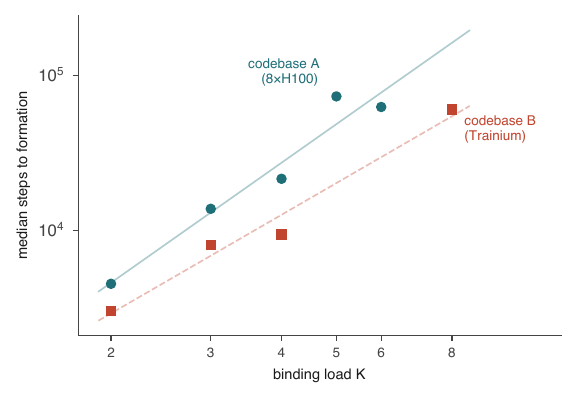}
\caption{\textbf{The cost-of-load law, in two codebases.} Median steps to formation against binding
load, both axes logarithmic, for two independent implementations; faint
lines are the committed power-law fits ($t \approx \TimeLawHCoef\,K^{\TimeLawHExp}$ and
$t \approx \TimeLawDCoef\,K^{\TimeLawDExp}$). The second corpus was analysed after the first fit was
fixed. The medians exclude runs that did not form within budget. They describe successful runs,
not an unconditional formation-time law; censoring does not provide a lower bound on the exponent.}
\label{fig:timelaw}
\end{figure}

 Direct task training has its own price in $K$. For
small transformers (1.3M--5M parameters) trained from scratch on the same task, median time to first
formation grows as a power of $K$. One implementation gives
$t \approx \TimeLawHCoef\,K^{\TimeLawHExp}$ ($R^2=\TimeLawHRsq$), and the independent second fit is
$t \approx \TimeLawDCoef\,K^{\TimeLawDExp}$ ($R^2=\TimeLawDRsq$). We fixed the first fit before examining the
second corpus. The functional form replicates, but the exponent does not resolve. The second corpus's $K{=}8$ cell holds two runs, and removing it moves the exponent to \TimeLawExpNoKeight. Both fits condition on formation within the training budgets; they do not estimate unconditional formation times.
Seed matters far beyond chance: across 24 seeds the variance of emergence rate is $\SeedOverdisp\times$ the binomial expectation, so per-seed comparisons must be paired. Figure~\ref{fig:timelaw} shows both fits; their conditional medians imply a factor of \TimeLawDblLo{}--\TimeLawDblHi{} in training steps when load doubles. Conditioning on success can alter the fitted exponent in either direction (Section~\ref{sec:censoring}).

The two scaling plots also have different response variables. Figure~\ref{fig:capacitylaw}
relates parameter count to zero-shot recall at inference, whereas
Figure~\ref{fig:timelaw} relates binding load to training steps among runs that formed.
Neither plot maps an additional training step to a change in a pretrained model's
$K_{50}$. Combining their fitted exponents would therefore not estimate the training
cost of raising zero-shot capacity.

\subsection{What conditioning on formation changes}\label{sec:censoring}
Let $T_K$ be the random time to first formation at load $K$, allowing $T_K=\infty$ for runs that
never form. Write $F_K(t)=\Pr(T_K\leq t)$. With a common observation budget $\tau$ within a load
cell and $F_K(\tau)>0$, the population distribution among completed runs satisfies
\begin{equation}
 \Pr(T_K\leq t\mid T_K\leq\tau)=\frac{F_K(t)}{F_K(\tau)},\qquad t\leq\tau.
 \label{eq:formation-conditioning}
\end{equation}
Define $F_K^{-1}(v)=\inf\{t:F_K(t)\geq v\}$. The lower median among completed runs is consequently
$m_{\mathrm{completed}}(K)=F_K^{-1}(F_K(\tau)/2)$, which is no larger than the unconditional
median $m_{\mathrm{full}}(K)=F_K^{-1}(1/2)$. If $F_K(\tau)<1/2$, the latter lies beyond the
budget and may be infinite. These are population statements under the stated budget assumption,
not confidence bounds for the reported sample medians.

Even when both medians are positive and finite, the ordering does not bound their scaling
exponents. Write $m_{\mathrm{completed}}(K)=w(K)m_{\mathrm{full}}(K)$ with $0<w(K)\leq1$.
For differentiable curves their log--log slopes differ by $d\log w/d\log K$, whose sign is
unrestricted by $w\leq1$. The two implementations therefore replicate a power-law description
among successful runs. Establishing an unconditional formation-time law would also require the
completion probabilities and censoring budgets, including the runs that did not form.

\section{Interference costs three separate things}\label{sec:robust}
Raw capacity and interference-robustness are distinct axes, and the distinction is sharper than a
single retained-capacity number suggests. A distractor block can damage a recall curve in three
ways: it can lower the ceiling the curve starts from, move the load at which the curve crosses its
midpoint, and change how steeply the curve falls. On our curves the three behave differently.

\paragraph{The ceiling falls, and its association with capacity is inconclusive.} Interference damages
recall at $K=1$, before any capacity question arises. Across the \IntCeilN{} continuous-curve models
the $\Dblock$-token block costs a median $\IntCeilMed$ of the $K{=}1$ ceiling, from $\IntCeilLo$
(Qwen2.5-1.5B) to $\IntCeilHi$ (Pythia-410M). The association with capacity is inconclusive (Spearman
$\rho=\IntCeilRho$, $p=\IntCeilP$). These curves do not establish either protection from greater
capacity or independence between capacity and ceiling loss.

\paragraph{The midpoint shifts, and the size of the shift depends on the normalisation.} Measured
against the uninterfered ceiling, the median retained fraction is $\TrnIntRetained$, with a
reduction in $\TrnIntReduced$ of $\TrnIntN$ models: interference roughly halves effective capacity.
Measured against each curve's own ceiling, the median retained fraction is $\IntOwnRetained$ with a
reduction in $\IntOwnReduced$ of $\IntOwnN$ models, so the median is close to unchanged. We report both,
because the disagreement is itself the finding rather than a choice between estimators. The two
differ by exactly the ceiling term above: own-ceiling normalisation divides each curve by a ceiling
that interference has already lowered, which removes the damage from the measurement. Pythia-410M is
the clearest case, losing $\IntCeilHi$ of its single-binding ceiling before the capacity axis is
consulted at all. We therefore take the uninterfered ceiling as the reference, and report the
own-ceiling figure so that the choice is visible. The fixed-reference association between capacity
and retained capacity is also inconclusive (Spearman $\rho=\TrnIntRho$, $p=\TrnIntP$, $n=\TrnIntN$).

\paragraph{The slopes show no consistent direction of change.} The logistic slope at $D=\Dblock$ divided by the slope at $D=0$ has median
$\IntSlopeRatio$ over the \IntSlopeN{} models with an uncensored fit at both loads, ranging
$\IntSlopeLo$ to $\IntSlopeHi$, with \IntSlopeSteeper{} of \IntSlopeN{} steeper. These summaries do
not establish equivalence of curve shape.

\begin{figure}[t]
  \centering
  \includegraphics[width=0.8\linewidth]{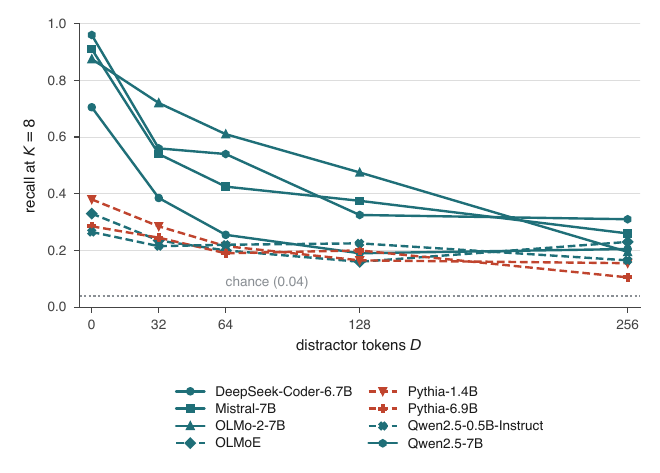}
  \caption{\textbf{Capacity $\neq$ robustness.} Recall vs.\ interference load $D$ at fixed $K=\RobustK$ for the
  eight models with committed curves (teal: modern recipe; clay: pre-2023). Dashed curves mark the
  \RobustLowKN{} models whose $\kstar$ is below $K=\RobustK$; they start at $\RobustDzeroLo$--$\RobustLowKDzeroHi$
  before any distractor arrives. Collapse rates do not follow the capacity ranking, so robustness is a second
  axis, measured separately.}
  \label{fig:robust}
\end{figure}

\paragraph{The $D$-sweep at fixed $K$.} We sweep $D$ at $K=\RobustK$ for the eight models with
committed curves (Figure~\ref{fig:robust}). At $D=0$, recall spans $\RobustDzeroLo$ to
$\RobustDzeroHi$ with median $\RobustDzeroMed$. The \RobustLowKN{} models with $\kstar$ below
$\RobustK$ start at $\RobustDzeroLo$--$\RobustLowKDzeroHi$, while the four models with
$\kstar\geq16$ start at $\RobustHighKDzeroLo$--$\RobustDzeroHi$. The 0.5B entry is the Instruct
variant of a base model with $\kstar=3$. By $D=\RobustDmax$ every model sits at
$\RobustDendLo$--$\RobustDendHi$ (median $\RobustDendMed$), and the collapse \emph{rates} do not
follow the capacity ranking. Our $D$-sweep parallels the interference-induced retrieval decline of \citet{pillm2506},
under a different manipulation: they overwrite earlier values for the same keys, whereas our block
is unrelated filler.

\subsection{What the two normalisations identify}\label{sec:normalisation}
The disagreement between normalisations has an exact interpretation before curve fitting. Write
$p_D(K)$ for recall at load $K$ and interference $D$, $b$ for the common chance level, and
$C_D=p_D(1)$ for single-binding recall. Suppose $C_0>b$ and $C_D>b$. Define
\begin{equation}
 r_D(K)=\frac{p_D(K)-b}{C_D-b},\qquad
 g_D=\frac{C_D-b}{C_0-b},\qquad
 \widetilde r_D(K)=\frac{p_D(K)-b}{C_0-b}=g_D\,r_D(K).
 \label{eq:ceiling-decomposition}
\end{equation}
Here $g_D$ measures retained single-binding performance above chance, while $r_D$ measures the
remaining load dependence relative to that performance. The last equality follows by cancelling
$C_D-b$. It is an identity for normalised recall, not for the ratio of fitted capacity estimates.

\paragraph{Midpoint consequence.} Treat load as continuous for the crossing definition. Suppose
$r_D$ is continuous and strictly decreasing on $K\geq1$,
and both crossings exist. For $1/2<g_D\leq1$, their locations satisfy
\begin{equation}
 K_{50}^{\mathrm{own}}(D)=r_D^{-1}(1/2),\qquad
 K_{50}^{\mathrm{fixed}}(D)=r_D^{-1}\!\left(\frac{1}{2g_D}\right)
 \leq K_{50}^{\mathrm{own}}(D).
 \label{eq:ceiling-crossing}
\end{equation}
The inequality follows because the inverse of a decreasing function is decreasing. If $g_D<1/2$,
the fixed reference criterion already fails at $K=1$; equality holds there when $g_D=1/2$.
These are equality crossings of a continuous curve, distinct from the first strictly-below grid
crossing used for $\kstar$.

This gives a concrete alternative to interpreting every capacity reduction as a loss of binding
storage. If interference lowers $g_D$ within $(1/2,1)$ while leaving $r_D=r_0$, the own-ceiling midpoint stays fixed
but the fixed-ceiling midpoint falls. The measurements above are compatible with this possibility;
they do not establish that it is the sole cause. Conversely, matching own-ceiling midpoints alone
does not establish $r_D=r_0$. The full profiles must agree. Reporting the ceiling, midpoint, and
slope therefore separates observable changes without assigning them to distinct neural mechanisms.
Section~\ref{sec:thresholds} gives the limits on threshold transfer; Section~\ref{sec:identification} treats the mechanistic interpretation.

\section{What geometry can explain}\label{sec:geometry-evidence}
The recall curves establish a behavioral limit at the query interface. Explaining that limit
requires a different kind of evidence. We examine the existing geometric comparisons and the
trial-level error analysis separately, because a cross-model association and an explanation of
individual retrieval failures are different claims.

The released geometric quantities are an effective dimension $\ddecl$ (a participation ratio)
and the mean off-diagonal squared cosine between entity vectors. Their provenance limits the
comparison: the plotting code reads a committed table, but the measurement script was not preserved,
so we cannot independently verify the token positions used for that table. Appendix~\ref{app:geometry}
reports the position and activation-scale artifacts in the available measurements. We retain
this limitation beside the results; these comparisons do not identify binding storage.

\subsection{Recall alone does not identify storage failure}\label{sec:identification}
Even a complete recall curve need not identify where retrieval fails. Consider a latent model in
which the queried value is retained with probability $s(K,D)$ and is accessed correctly, conditional
on retention, with probability $a(K,D)$. If either stage fails, the system guesses uniformly from
the obligation pool. Its recall is
\begin{equation}
 p_D(K)=b+(1-b)s(K,D)a(K,D),\qquad 0\leq b<1.
 \label{eq:storage-access}
\end{equation}
For any curve with $b\leq p_D(K)\leq1$, set $u(K,D)=[p_D(K)-b]/(1-b)$.
Both $(s,a)=(u,1)$ and $(s,a)=(1,u)$ generate exactly that curve. Wherever $p_D(K)<1$, these
describe different allocations of failure between retention and access. Hence recall does not
identify those factors without additional constraints. The construction does not assume that
transformers use these two stages; it demonstrates ambiguity in interpreting behaviour alone.
Activation interventions can provide additional evidence, as in \citet{binding2310,mixing2510}.
The present cross-model geometry and recall curves do not supply that identification.

\subsection{Recipe contrasts and capacity prediction are different tests}\label{sec:geometry}
\begin{figure}[t]
  \centering
  \includegraphics[width=\linewidth]{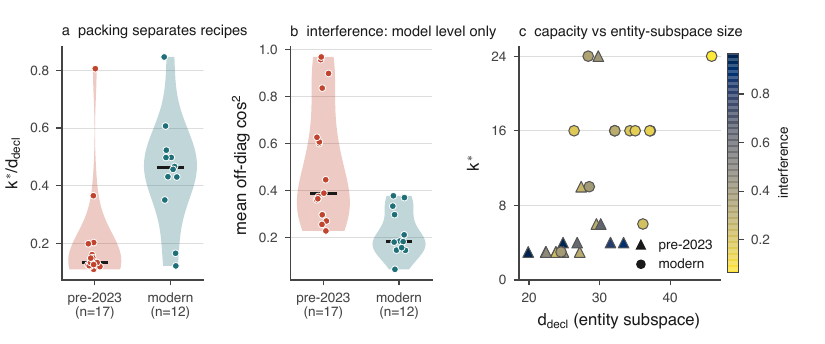}
  \caption{\textbf{How a model spends directions on the entities it must keep distinct, at the model
  level.} \textbf{(a)} Packing efficiency
  $\kstar/\ddecl$ of the declaration-site \emph{entity} subspace and \textbf{(b)} pairwise interference
  separate the recipes at the model level (each point is a model; black dash: median). \textbf{(c)} $\kstar$
  against $\ddecl$ over the \NModels{} measurable models, colored by interference. Both group comparisons are
  significant at the model level, while the family-clustered recipe comparisons are reported in Table~\ref{tab:famclust}.
  These panels do not establish within-family prediction or a retrieval mechanism.}
  \label{fig:recipe}
\end{figure}

Capacity has two geometric correlates at the model level (Figure~\ref{fig:recipe}). Across the \NModels{} models, $\kstar$ rises with $\ddecl$
(Pearson $r=\DdeclPearson$, Spearman $\rho=\DdeclSpearman$; partial $\DdeclPartial$ controlling for interference) and falls with
pairwise interference (Pearson $r=\OffdiagPearson$; partial $\OffdiagPartial$). Modern recipes show a higher packing efficiency
$\kstar/\ddecl$ (median $\PackModMed$ vs.\ $\PackOldMed$) at lower interference ($\OffdiagModMed$ vs.\ $\OffdiagOldMed$), and every one of these
comparisons is significant at the model level. A simple account is that older recipes place entity codes on
more overlapping directions, while modern recipes spread them. This would connect binding capacity to the
superposition regime of \citet{superposition2505}, where loss scales inversely with model dimension under
strong superposition and weight decay controls superposition in their toy model. It would also sit
alongside \citet{largercap2605}, who attribute larger models' advantage to reduced gradient interference
on rare tasks.

\paragraph{Thirty models are not thirty independent samples.} They are \emph{ten families}, and Pythia and
OLMo contribute six apiece. A model-level test treats siblings as independent draws. Under a bootstrap clustered on families, and under a family-level test, the recipe contrasts have different uncertainty (Table~\ref{tab:famclust}).

\begin{table}[h]
\centering
\caption{Family-clustered comparisons. Medians by recipe with the family-clustered bootstrap
interval and the family-level $p$-value for each quantity. Bold highlights capacity and its
derived ratio. These are recipe contrasts, not tests of within-family capacity prediction.}
\label{tab:famclust}
\small
\begin{tabular}{@{}lrrcc@{}}
\toprule
quantity & old med. & modern med. & family-clustered $95\%$ CI & family $p$ \\
\midrule
capacity $\kstar$                   & $\KstarOldMed$  & $\KstarModMed$ & $\KstarFamCI$ & $\mathbf{\KstarFamP}$ \\
packing efficiency $\kstar/\ddecl$  & $\PackOldMed$ & $\PackModMed$ & $\PackFamCI$ & $\mathbf{\PackFamP}$ \\
subspace dimension $\ddecl$         & $\DdimOldMed$ & $\DdimModMed$ & $\DdimFamCI$ & $\DbindFamP$ \\
interference                        & $\OffdiagOldMed$ & $\OffdiagModMed$ & $\OffdiagFamCITab$ & $\OffdiagFamP$ \\
\bottomrule
\end{tabular}
\end{table}

The clustered interval for $\ddecl$ includes zero and its family-level test is not significant.
For interference, the interval includes zero while the family-level test gives a different
summary of uncertainty. These procedures test differences between recipes; neither estimates
whether geometry predicts capacity within a family. A null recipe contrast would not exclude
such a relation. Packing efficiency contains $\kstar$ in its numerator and therefore cannot
serve as an independent predictor of that same outcome.

The capacity contrast remains the result of Section~\ref{sec:capacity}. Establishing a geometric
account would require family-adjusted prediction using geometry measured independently of the
capacity target. The present table does not supply that test. We report the model-level patterns
alongside the clustered recipe comparisons rather than infer a mechanism from either. With ten
families (six old, four modern), family-level inference also has fewer independent units than
the model count suggests.

\subsection{Cross-model capacity and within-model retrieval separate}\label{sec:notmech}
Capacity across models and retrieval within a model sit at different explanatory levels, and the
geometry above speaks only to the first.
The packing story could be read as a competitor to the positional-mechanism account of
\citet{mixing2510}, who attribute multi-binding retrieval failure to a positional circuit that grows noisy as
bindings accumulate. We ran the comparison on our own trials, regressing per-trial error on serial position
and on the pairwise overlap of the declaration-site entity vectors. \textbf{The geometry loses.} Once position is controlled,
subspace overlap adds between $\HtHIncLo$ and $\HtHIncHi$ AUROC on all seven models tested; position alone reaches
$\HtHPosAurocLo$--$\HtHPosAurocHi$.

Neither our geometry nor a bare positional pointer predicts \emph{which} distractor is emitted in our
setting: the emitted obligation is the maximum-overlap competitor on $\HtHFoilOvlLo$--$\HtHFoilOvlHi$ of failures against a
$\HtHFoilOvlChance$ chance rate, and a positional neighbour at or below chance. We note that \citet{mixing2510} do predict
the emitted token, with a causal model that combines the positional mechanism with a \emph{lexical} and a
\emph{reflexive} one and reaches $95\%$ agreement with next-token distributions; our head-to-head pits only their positional term against our overlap term, and it is the positional term alone that wins.

The ratio $\kstar/\ddecl$ is a cross-model description that includes the capacity target.
The trial-level comparison asks a different question: whether overlap improves prediction beyond
position. Its reported increments do not establish a useful improvement.

One caveat on the comparison itself: our error curve is a primacy curve rather than the U-shape of \citet{mixing2510}, and includes a $\Dblock$-token
interference block between the declarations and query. The block is a possible explanation for
the missing recency effect, but this comparison does not isolate its cause. The two settings are not
directly comparable.

\subsection{An earlier entity state can still serve as an address}\label{sec:address}
An entity token before its assigned value cannot directly represent a change confined to that
later value. This constraint concerns the information available at a site, not every possible
role the site can play in a distributed computation.

This causal-order argument has a narrower implication than excluding the site from binding.
Under deterministic causal attention, the state at token $i$ is a function of the tokenised prefix
$x_{\leq i}$. Replacing the obligation in the suffix while keeping that prefix, positions, and mask
fixed leaves the entity state unchanged. It cannot directly encode the newly supplied value, but
it can encode an address that a later state associates with that value. Such a distributed role is
consistent with the shared binding IDs of \citet{binding2310}. Failure to decode the obligation
from the entity state therefore does not rule out a causal role in retrieval. It rules out using
that decoder as evidence that the assigned value is already represented there.

\section{Using capacity comparisons}\label{sec:use}
A model selected for a binding task must meet a recall requirement at the relevant interference
load. An own-ceiling midpoint instead describes how much performance remains relative to that
model's single-binding baseline. Section~\ref{sec:thresholds} shows that even rankings under
relative criteria need not transfer across thresholds. Absolute recall imposes another condition.

\subsection{Absolute recall defines the feasible loads}
Let $\pi\in(b,1)$ be a required recall probability and assume $C_D>b$. The empirical normalisation
in Equation~\eqref{eq:ceiling-decomposition} gives the exact equivalence
\begin{equation}
 p_D(K)\geq\pi
 \quad\Longleftrightarrow\quad
 r_D(K)\geq\frac{\pi-b}{C_D-b}.
 \label{eq:absolute-recall}
\end{equation}
Thus a common absolute requirement generally corresponds to a different retained-performance
criterion for each model and interference condition. If recall is nonincreasing for $K\geq1$
and $C_D<\pi$, no load meets the requirement. If $C_D=\pi$, load one meets it, as can a plateau;
if $C_D>\pi$, the criterion lies between zero and one, but its crossing may remain outside the
measured range. These statements concern the underlying curves. Finite-sample estimates require
uncertainty on both the ceiling and the recall probabilities.

For a measured set of load--interference pairs $\mathcal G$, report the feasible set
$\{(K,D)\in\mathcal G:p_D(K)\geq\pi\}$ with its uncertainty. This preserves the task requirement
without converting unmeasured loads into fitted capacity claims. It complements the midpoint,
which remains useful for comparing the shape of the decline relative to baseline performance.

\subsection{Analyses that distinguish the remaining explanations}
The equations motivate tests that can reuse saved trial outputs. Threshold transfer should compare
crossings at fixed recall criteria on the same model set, then report the additional selection
caused by censoring. Direct empirical crossings provide a check on logistic extrapolation. A
reversal supported within the measured range would reject a threshold-independent model ranking;
a reversal outside that range would remain a model prediction.

The interference comparison admits a more specific test than a retained-capacity ratio. Under a
ceiling-only account, the own-ceiling profile is unchanged, so the interfered recall predicted
from the baseline profile is $b+(C_D-b)r_0(K)$. Systematic residuals across load would reject this
account. Ceiling estimation and profile evaluation should use separate trials or propagate their
shared uncertainty. Agreement within a predeclared tolerance would support this restricted account
on the measured grid, without identifying a neural mechanism.

The geometric comparison should ask whether independently measured geometry improves prediction
beyond model size and family, with validation that holds out families. The trial-level analogue
compares position alone with position plus overlap on held-out contexts. A positive result at
one level would not establish the other. Formation-time analyses should retain unsuccessful runs
and their observation budgets, reporting completion probability alongside time among completed
runs. These are proposed analyses, not additional results of this paper.

\section{Working memory, interpretability, and alignment}\label{sec:implications}
Binding capacity matters when correct behavior depends on retaining which fact or rule belongs
to which entity. The connection to working memory is operational: our task measures temporary
access to relations supplied in context. The connection to alignment requires another step,
from recovering a relation to using it in an action. The following deductions make that step
explicit; they do not report an alignment evaluation.

\subsection{Working memory is a task-dependent ability}
An input window specifies how much text can be supplied. Our capacity statistic specifies how
many bindings can be recalled at a chosen criterion. The two quantities answer different
questions. \citet{lost2307} show that the position of relevant information affects long-context
retrieval, while \citet{pillm2506} study interference from overwritten values. Our unrelated
filler manipulation does not measure the same form of updating. Working memory therefore
provides a useful family of tasks, rather than a single capacity shared by these measurements.

Increasing $K$ in our task also changes the declaration sequence. The present size law does not
isolate binding count from every change in token length or query distance. A matched-length
comparison would vary the number of distinct relations while holding total tokens and the
queried entity's position fixed. A separate manipulation would change whether a value is retained,
overwritten, or composed with another value. Such tests would determine which operations the
capacity statistic predicts. They would not establish a human-like memory architecture merely
because a behavioral curve resembles a working-memory limit.

\subsection{From binding to world-state tracking}
To test whether binding supports a model's approximation of a world state, one must ask more
than whether it retrieves a declared fact. A stronger test would require the model to revise an
entity's properties after an event and use the revised state to predict or choose an action.
Our trials contain neither state changes nor action outcomes:
each entity receives one obligation, and the model answers one query. Controlled game sequences
have supported tests of internal board-state representations and their causal use
\citep{li2023othello}; state-maintaining prompts have been used to track changes during embodied
planning \citep{yoneda2024statler}. These studies motivate the next operations to measure, not an
inference that our capacity curve measures world-model quality.

Othello makes the comparison precise. A board square has a changing value, while a legal move
depends on the configuration of several squares. The analogy is entity-specific state, not a
numerical equivalence between our load $K$ and board size. Equation~\eqref{eq:storage-access}
shows that the same recall curve can arise when a value is unavailable or when it is retained
but inaccessible at the query. Thus poor recall does not show that the state is absent internally.
Equation~\eqref{eq:joint-recall} shows that random single-query accuracy does not determine the
chance of retrieving every property needed for one decision. Thus good midpoint capacity does
not establish a coherent multi-entity state. Even perfect static lookup could solve every
present trial without learning a state-transition rule.

Othello-GPT goes beyond output accuracy by probing board state and intervening on the decoded
representation before measuring changed move predictions \citep{li2023othello}. In our setting,
the declaration-site entity vector precedes the assigned value (Section~\ref{sec:address});
a probe of the current value should instead use a later site that can access it. An analogous
binding intervention would change one entity's represented value, predict a selective answer
change, and check that other bindings remain unchanged. Even that result would concern static
retrieval until state updates and transitions are tested.

A direct extension would compare static recall, overwritten values, and action-conditioned state
transitions on held-out entity and value assignments. It would score current-state retrieval and
the resulting prediction or action separately. Matched-length and matched-position controls are
needed before attributing differences to binding load rather than extra tokens or query distance.
One can then ask whether an absolute-recall criterion at a stated $K$ and $D$ predicts performance
on the later operations. A correlation across model families would remain confounded by model
size and training recipe; a controlled training comparison would give a clearer test of transfer.

The transfer question has two parts. Synthetic numerical key--value training improved
multi-document question answering in the models tested by \citet{xiong2024needles}.
In contrast, \citet{yen2025helmet} find that synthetic needle-in-a-haystack scores do not
reliably predict performance across downstream long-context tasks. These findings test
different claims: an effect of training in one setting and the predictive value of a
benchmark across settings. Neither establishes that our measured $K_{50}$ predicts a
downstream task or that training to raise it improves that task. The matched
base-versus-code-trained comparisons of \citet{kim2405} further show that training
data can change entity-tracking performance, without identifying binding capacity as
the transferred ability.

Training to raise the relative $K_{50}$ is not itself the target. Because $K_{50}$ is defined
against each model's own single-binding ceiling, an increase in $K_{50}$ need not improve
absolute recall at the load and interference level a task requires (Section~\ref{sec:use}).
A predictive test should ask whether absolute recall at fixed load and interference
predicts held-out task performance beyond a size-only baseline, reporting within-family
and held-out-family results separately.
A binding curriculum should be compared with equal-token, equal-compute training controls,
using held-out bindings and downstream tasks to assess gains and losses. Relevant endpoints
include dialogue state tracking, program execution, multi-document question answering, and
state-dependent planning. General language performance and policy outcomes need separate
measurement rather than inference from recall.

\subsection{Single-query recall does not determine joint reliability}
Consider an episode that requires correct retrieval of $m\geq1$ bindings. Let $R_j$ be the event
that binding $j$ is retrieved correctly and let $p_j=\Pr(R_j)$. The events must refer to the
same episode distribution and a specified retrieval protocol. For $R=\bigcap_{j=1}^m R_j$,
\begin{equation}
 \max\!\left\{0,1-\sum_{j=1}^{m}(1-p_j)\right\}
 \leq \Pr(R) \leq \min_j p_j.
 \label{eq:joint-recall}
\end{equation}
The lower bound follows by applying the union bound to the failure events $R_j^c$; the upper
bound follows from $R\subseteq R_j$. No independence assumption is needed. A uniform query,
chosen independently of episode outcomes, measures the mean marginal recall
$\bar p=m^{-1}\sum_j p_j$. The lower bound is then $\max\{0,1-m(1-\bar p)\}$.
The same average can accompany different dependencies among errors and therefore different
probabilities of retrieving every binding. Independent events with equal marginals would give
$\Pr(R)=\bar p^{\,m}$; these assumptions are not established here.

For a tolerated joint failure probability $\epsilon\in(0,1)$, a sufficient condition is
$\sum_j(1-p_j)\leq\epsilon$. In particular, marginal guarantees
$p_j\geq1-\epsilon/m$ for every required binding suffice. This is a distribution-free
sufficient condition, not a necessary one or a guarantee estimated from our current trials.

This distinction matters when several recovered constraints jointly determine an action. A
midpoint summary does not establish reliable joint retrieval, and a weak lower bound does not
prove that joint retrieval is poor. The paper's random-query trials do not directly measure
success on a multi-action episode. Estimating that quantity requires a protocol that defines
all required retrievals jointly, including any effects of earlier generated answers.

\subsection{Correct retrieval and policy adherence are separate outcomes}
As a controlled example, an instruction-following task can assign each resource an authorized
action and a source of authority. A model can retrieve the action correctly while applying the
wrong authority, or retrieve the wrong action despite respecting the source hierarchy.
\citet{hierarchy2404} study instruction priority explicitly. Our entity--obligation task has
neither competing authority levels nor a policy-adherence endpoint, so it cannot measure that
ability. This distinction also prevents interpreting the unchanged post-training capacity
threshold in Section~\ref{sec:capacity} as evidence that alignment training has no effect.

Let $V$ denote a violation in a specified policy task and let $R$ denote correct retrieval of
all bindings required by that task. Write $r=\Pr(R)$, $e_1=\Pr(V\mid R)$, and
$e_0=\Pr(V\mid R^c)$, with $0<r<1$. The law of total probability gives
\begin{equation}
 \Pr(V)=r e_1+(1-r)e_0.
 \label{eq:policy-error}
\end{equation}
Thus retrieval reliability alone does not identify the violation rate. In a comparison that
holds both conditional violation rates fixed, a change from $r$ to $r'$ gives
$\Pr'(V)-\Pr(V)=(r'-r)(e_1-e_0)$. Better retrieval reduces violations under this assumption
only when $e_1<e_0$. A retrieval intervention can also change those conditional rates, and
conditioning on naturally successful retrieval can select easier episodes. The identity is
therefore an accounting relation, not a causal mediation result. At $r=0$ or $r=1$, use the
unconditional joint probabilities because one conditional rate is undefined.

A testable extension would randomize a retrieval aid against a matched control on the same
synthetic policy tasks. It would score retrieved bindings and final actions separately, and
include permitted requests to detect indiscriminate refusal. If the aid improves retrieval
without reducing violations, the proposed benefit to adherence fails for that intervention.
Even a positive randomized effect would establish the effect of the aid; attributing it
specifically to retrieval would require further mediation assumptions. This is a bounded
connection to alignment research, not a claim that greater binding capacity makes a model aligned.

\subsection{Interpretability must distinguish representation from use}
Section~\ref{sec:identification} shows why a recall curve cannot distinguish retention from
access. Section~\ref{sec:address} shows why the absence of a newly assigned value at an earlier
entity token does not exclude an address role. Together, they constrain what an interpretation
experiment must establish. A decoder tests whether its chosen representation exposes a label;
it does not by itself test whether the model uses that information to choose its answer.

The intervention designs of \citet{li2023othello,binding2310,mixing2510} provide examples of
testing use through targeted changes in activations. For our setting, a proposed test would swap an entity
address or a later value representation while preserving the task's other bindings. It should
predict which answer changes, not merely whether accuracy falls. Unrelated donors and matched
perturbations would test whether the change reflects binding content rather than generic
disruption. Such an intervention would connect the capacity curve to a candidate mechanism;
the present geometry correlations and probes do not supply that connection.

\section{Related work}\label{sec:related}
\citet{kim2405} show with matched base and code-trained pairs that code pretraining drives entity tracking,
while alignment adds no consistent benefit. We define the capacity threshold $\kstar$ and sweep it across
$\NModelsRaw$ models, which exposes a flat Pythia ladder beside an $\KstarRangeFactor\times$ recipe span.
\citet{binding2310} identify a binding-ID mechanism whose fidelity improves with scale. Mechanism fidelity and
capacity ceiling may vary separately, a distinction that calls for a direct dissociation experiment.
Architecture also changes entity-tracking performance in closely matched transformer and hybrid models \citep{archentity2606}, so our use
of \emph{recipe} includes architecture. Across model families, the same composed task can rely on different
circuits \citep{mechhet2606}, which motivates separating cross-model geometry from trial-level mechanism tests
(Section~\ref{sec:geometry-evidence}). On the scaling side,
\citet{kaplan2020} and \citet{hoffmann2022} fit loss against scale; the $K_{50}$ law prices a specific
in-context capacity instead, and its regime boundary shows what such zero-shot laws are laws of.
\citet{muennighoff2023} extend the loss accounting to data-constrained training, pricing the decaying
value of a repeated token; our boundary prices a named capability rather than loss.
\citet{allenzhu2024} price parametric knowledge in bits per parameter; in-context binding state is the
complementary quantity, held in activations rather than weights. Related mechanisms, induction heads and
associative memories, are described in \citet{olsson2022} and \citet{bietti2023}.

Cognitive framings of the same limit measure it differently. \citet{gong2024} adapt the $n$-back
task to assess verbal and spatial working memory; \citet{xiong2604} report load-dependent
decline and human-like interference signatures across pretrained models. Our analysis instead
estimates a load-capacity scaling law across model sizes. On the applied side, \citet{ruler2404} include
multi-key retrieval against hard distractors and a variable-tracking task in a long-context
benchmark. Our task varies binding count and distractor length on separate axes and fits capacity
across models; it does not fully separate binding count from total token length.

Work on world-state representations asks whether a model encodes and uses a changing state.
\citet{li2023othello} use a controlled board game and activation interventions, while
\citet{yoneda2024statler} maintain an explicit state estimate for planning. Our static binding
task isolates an earlier operation and does not test those claims. \citet{xiong2024needles}
report transfer from synthetic retrieval training to multi-document question answering, which
motivates but does not settle the transfer question for our capacity measure.
\citet{yen2025helmet} find that synthetic needle retrieval scores do not reliably predict
downstream long-context performance. Whether our load-dependent measure predicts performance
beyond model size remains a separate empirical question.

\section{Discussion and conclusion}\label{sec:limits}
\paragraph{Limitations.}
The recipe split is observational. The \NModels{} measurable models differ in data, tokenizer, optimizer,
architecture, and era. Our comparison identifies a recipe-level regularity without causal attribution. The $\kstar$ statistic is also right-censored at $\KstarMax$ by the entity pool, making the low end of
the range more informative than the ceiling.

The continuous curves give a different view at or below 3B: recipe adds no detectable effect after scale is
controlled. Curves above 3B are needed to determine whether the broad recipe split returns at larger scale.
The $K_{50}$ scaling fit was found post hoc and awaits pre-registered replication.
The synthetic, single-token task provides direct control over load and interference. The resulting laws apply
to this task, these model ranges, and these estimators. The geometric quantities in Section~\ref{sec:geometry-evidence}
describe cross-model correlates and do not identify a trial-level mechanism.
The sensitivity checks in Sections~\ref{sec:sensitivity} and~\ref{sec:recipe-sensitivity}, and the
interference decomposition in Section~\ref{sec:robust}, include exploratory re-analyses of the
existing corpus. These analyses were not predicted in advance. The threshold and identification
results are conditional deductions, and the tests in Section~\ref{sec:use} remain proposals.

\paragraph{Conclusion.}
The broad sweep and continuous curves measure different parts of in-context binding capacity. On continuous
curves at or below 3B, $K_{50}$ follows a size law and the recipe coefficient spans zero after scale is
controlled; across models to 12B, $\kstar$ shows recipe-associated differences across scale.

Direct task training exceeds an extrapolation of the zero-shot law by $\BoundFactor\times$, under
different measurement criteria. Interference robustness varies on another axis. The normalisation
identity explains why a loss at the single-binding ceiling can appear as a capacity loss, and the
threshold analysis specifies when a scaling exponent transfers to other recall criteria.
Capacity numbers should therefore state the task, instrument, training regime, and interference
load that produced them. These qualifications turn the capacity curve into a description of usable
recall while leaving the storage mechanism open.

\section*{Acknowledgements}
We would like to thank Kevin Li for his help with proofreading, for helping us to better write
the paper for readability, and for the suggestions that led to this paper.

\bibliographystyle{plainnat}

\clearpage
\appendix

\section{Measurement diagnostics for declaration-site geometry}\label{app:geometry}
This appendix reports measurement diagnostics for the declaration-site geometric quantities.
Section~\ref{sec:geometry-evidence} presents the recipe comparisons and trial-level prediction tests.

\paragraph{The geometric quantities, and what they measure.} At the declaration site we take one residual
vector per declared entity and compute $\ddecl$, the participation ratio of those vectors (their effective
dimension), and the mean off-diagonal squared cosine among them (pairwise \emph{interference}).

The natural name for these quantities, a binding geometry, overstates them. Each declaration reads
\texttt{entity: obligation.}, so the entity token \emph{precedes} its obligation, and attention is causal. The
hidden state at an entity token cannot directly encode a newly assigned obligation that appears later.
Section~\ref{sec:address} distinguishes this constraint from a possible address role. With the per-slot mean removed, the entity decodes at $1.000$ from the
entity-token residual. Its bound obligation decodes at $0.017$--$0.054$ against a chance of $\Chance$
(Pythia-1.4B/6.9B, Qwen2.5-7B, OLMo-2-7B). The quantities $\ddecl$ and interference therefore describe how
many directions the declaration-site entity code uses and how much those directions overlap. Capacity can
depend on this geometry, but the geometry does not measure the binding itself.

\subsection{Measurement artifacts at the declaration site}\label{sec:confounds}
Each of these artifacts produces a wrong but plausible-looking number, and each is easy to walk into.

\paragraph{1. Position dominates the raw vectors.} Entity $i$ sits in slot $i$, so declaration-site hidden
states are overwhelmingly positional: the per-slot mean carries $\PosVarLoPct$--$\PosVarHiPct\%$ of their
variance on \GeoNNonOlmo{} of the \GeoNModels{} models with committed geometry; the exception is OLMo-2, at
$\PosVarOlmoPct\%$. A participation ratio taken on the raw vectors reports an effective dimension of at most
$\DrawHi$ and a pairwise squared cosine of $\RawIntLo$--$\RawIntHi$: it is describing \emph{position}.
Estimating the positional component as the per-slot mean across trials and measuring on the residual gives an
interference of $\ResidIntLo$--$\ResidIntHi$ on all \GeoNModels{} models: the entity code is close to
orthogonal to the positional one.

\paragraph{2. Massive activations dominate the covariance.} Prior work identifies massive activations
and their relation to attention sinks \citep{massive2402, sink2410}. In our measurements, after position
is removed, a \emph{covariance} participation ratio is dominated by these few activation directions.
It reports an effective dimension of $\DcovQsevenB$ for Qwen2.5-7B, a model that carries $\TopOneQsevenB\%$ of its residual variance
in a single coordinate. Standardizing per coordinate gives $\DdimQsevenB$, and the dimension then scales sensibly with
model size within the family ($\DdimQonefiveB$, $\DdimQsevenB$, $\DdimQfourteenB$ for $1.5$B, $7$B, $14$B).

Both artifacts are \emph{installed by pretraining}. Across Pythia-1.4B checkpoints the positional variance
fraction rises from $\PosVarDevFirst$ to $\PosVarDevLast$ and the top coordinate's share from $\TopOneDevFirst$ to $\TopOneDevLast$, so a measurement
that behaves well on an early checkpoint can invert on a converged one. Under the corrected estimator, the
declaration-site subspace expands and then plateaus. The covariance ratio instead suggests compression.

\paragraph{3. The site precedes the assigned value.} The declaration-site vector is taken at the entity token, which causally
precedes its obligation, and the probes above decode the entity at $1.000$ and the obligation at chance.
Thus $\ddecl$ is the effective dimension of the entity code, and
packing efficiency measures how the model spends directions on entities that it must hold apart.

No downstream figure exposes this. The artifact is visible only in the projection of the
declaration-site residuals onto the obligation classes, which gives no separation in any frame, including
a supervised one. A
further provenance caveat: the plotting generator for Figure~\ref{fig:recipe} reads a
committed table, but the \emph{measurement} script that produced the $\ddecl$ values was not preserved, so we
cannot independently verify the token position at which those particular numbers were taken. Re-deriving them at
a position that \emph{can} carry a binding is the correct next measurement: the query site, where on \QueryDecodeN{} models the retrieved obligation decodes at
$\QueryDecodeLo$--$\QueryDecodeHi$ against a chance of $\Chance$.

\end{document}

%% file: numbers.tex
\newcommand{\KstarMin}{3}               
\newcommand{\KstarMax}{24}               
\newcommand{\KstarRangeFactor}{8}       
\newcommand{\NModels}{29}                
\newcommand{\KstarOldMed}{3}            
\newcommand{\KstarModMed}{16}            
\newcommand{\KstarFamCI}{[+10,\,+13]}             
\newcommand{\KstarFamP}{0.012}              
\newcommand{\PackFamP}{0.010}               
\newcommand{\DdimFamCI}{[-0.003,\,+11.77]}              
\newcommand{\DbindFamP}{0.114}              
\newcommand{\OffdiagFamP}{0.038}            
\newcommand{\KstarModelP}{4.8\times 10^{-4}}            
\newcommand{\NOldModels}{17}             
\newcommand{\NModernModels}{12}          
\newcommand{\PackOldMed}{0.134}             
\newcommand{\PackModMed}{0.462}             
\newcommand{\PackFamCI}{[+0.279,\,+0.473]}              
\newcommand{\DdimOldMed}{24.86}             
\newcommand{\DdimModMed}{33.23}             
\newcommand{\OffdiagOldMed}{0.388}          
\newcommand{\OffdiagModMed}{0.183}          
\newcommand{\OffdiagFamCITab}{[-0.629,\,+0.013]}        
\newcommand{\QwenLadder}{3\to10\to24\to16}             
\newcommand{\Dblock}{256}                 
\newcommand{\Chance}{0.040}                 
\newcommand{\DdeclPearson}{+0.64}           
\newcommand{\DdeclSpearman}{+0.77}          
\newcommand{\OffdiagPearson}{-0.52}         
\newcommand{\DdeclPartial}{+0.54}           
\newcommand{\OffdiagPartial}{-0.34}         
\newcommand{\QueryDecodeN}{5}           
\newcommand{\QueryDecodeLo}{0.51}          
\newcommand{\QueryDecodeHi}{0.86}          
\newcommand{\NModelsRaw}{30}             
\newcommand{\NOldRaw}{18}                
\newcommand{\NModernRaw}{12}             
\newcommand{\PythiaLadder}{3,4,6,4,4}           
\newcommand{\PythiaKstarLo}{3}          
\newcommand{\PythiaKstarHi}{6}          
\newcommand{\KstarOlmoE}{6}             
\newcommand{\KstarOptOneThree}{10}       
\newcommand{\KstarOlmoLadder}{16}        
\newcommand{\PythiaSeventyCeiling}{0.070}   
\newcommand{\RobustK}{8}                
\newcommand{\HtHPosAurocLo}{0.593}          
\newcommand{\HtHPosAurocHi}{0.843}          
\newcommand{\HtHIncLo}{-0.0005}               
\newcommand{\HtHIncHi}{+0.0048}               
\newcommand{\HtHFoilOvlLo}{0.12}           
\newcommand{\HtHFoilOvlHi}{0.19}           
\newcommand{\HtHFoilOvlChance}{0.20}       
\newcommand{\DcovQsevenB}{1.3}            
\newcommand{\TopOneQsevenB}{54}          
\newcommand{\DdimQonefiveB}{12.0}          
\newcommand{\DdimQsevenB}{35.3}            
\newcommand{\DdimQfourteenB}{81.1}         
\newcommand{\PosVarDevFirst}{0.15}         
\newcommand{\PosVarDevLast}{0.99}          
\newcommand{\TopOneDevFirst}{0.001}         
\newcommand{\TopOneDevLast}{0.17}          
\newcommand{\GeoNModels}{8}             
\newcommand{\GeoNNonOlmo}{7}            
\newcommand{\PosVarLoPct}{98.1}            
\newcommand{\PosVarHiPct}{99.9}            
\newcommand{\PosVarOlmoPct}{23}          
\newcommand{\RawIntLo}{0.84}               
\newcommand{\RawIntHi}{1.00}               
\newcommand{\DrawHi}{1.03}                 
\newcommand{\ResidIntLo}{0.006}             
\newcommand{\ResidIntHi}{0.009}             
\newcommand{\RobustLowKN}{4}            
\newcommand{\RobustLowKDzeroHi}{0.38}      
\newcommand{\RobustHighKDzeroLo}{0.70}     
\newcommand{\RobustDmax}{256}             
\newcommand{\RobustDzeroLo}{0.27}          
\newcommand{\RobustDzeroHi}{0.96}          
\newcommand{\RobustDzeroMed}{0.54}         
\newcommand{\RobustDendLo}{0.10}           
\newcommand{\RobustDendHi}{0.31}           
\newcommand{\RobustDendMed}{0.20}          
\newcommand{\CapAlpha}{0.820}               
\newcommand{\CapAlphaCILo}{0.674}           
\newcommand{\CapAlphaCIHi}{1.251}           
\newcommand{\CapRsq}{0.73}                 
\newcommand{\CapLofoLo}{0.76}              
\newcommand{\CapLofoHi}{1.09}              
\newcommand{\CapFitN}{11}                
\newcommand{\TrnRecipeCoef}{+0.123}          
\newcommand{\TrnRecipeCILo}{-0.521}          
\newcommand{\TrnRecipeCIHi}{0.363}          
\newcommand{\TrnScaleCoef}{+0.818}           
\newcommand{\TrnScaleCILo}{0.697}           
\newcommand{\TrnScaleCIHi}{1.206}           
\newcommand{\TrnOptResolved}{38.1}         
\newcommand{\TrnIntRetained}{0.535}         
\newcommand{\TrnIntReduced}{10}          
\newcommand{\TrnIntN}{11}                
\newcommand{\TrnIntRho}{+0.20}              
\newcommand{\TrnIntP}{0.56}                
\newcommand{\BoundMeasured}{3.7}          
\newcommand{\BoundPredicted}{0.14}         
\newcommand{\BoundFactor}{26}            
\newcommand{\CapLooMax}{0.11}              
\newcommand{\TimeLawHCoef}{766}           
\newcommand{\TimeLawHExp}{2.58}            
\newcommand{\TimeLawHRsq}{0.95}            
\newcommand{\TimeLawDCoef}{667}           
\newcommand{\TimeLawDExp}{2.12}            
\newcommand{\TimeLawDRsq}{0.97}            
\newcommand{\TimeLawExpNoKeight}{1.7}     
\newcommand{\TimeLawDblLo}{4}           
\newcommand{\TimeLawDblHi}{6}           
\newcommand{\SeedOverdisp}{9.05}           

%% file: numbers_addenda.tex
\newcommand{\SlopeLo}{0.499}                
\newcommand{\SlopeHi}{1.801}                
\newcommand{\SlopeN}{11}                    
\newcommand{\SlopeCorrK}{-0.837}            
\newcommand{\SlopePartialN}{-0.169}         
\newcommand{\AucExp}{+0.329}                
\newcommand{\AucRsq}{0.799}                 
\newcommand{\AucN}{12}                      
\newcommand{\AucCILo}{+0.286}               
\newcommand{\AucCIHi}{+0.480}               
\newcommand{\AucExpUnc}{+0.304}             
\newcommand{\AucRsqUnc}{0.821}              
\newcommand{\CeilLo}{0.918}                 
\newcommand{\CeilHi}{0.99}                  
\newcommand{\CeilNAbove}{11}                
\newcommand{\CeilN}{12}                     
\newcommand{\CeilOutlier}{0.589}            
\newcommand{\CeilRsq}{0.259}                
\newcommand{\IntCeilMed}{0.137}             
\newcommand{\IntCeilLo}{0.016}              
\newcommand{\IntCeilHi}{0.538}              
\newcommand{\IntCeilN}{12}                  
\newcommand{\IntCeilRho}{-0.473}            
\newcommand{\IntCeilP}{0.142}               
\newcommand{\IntOwnRetained}{0.963}         
\newcommand{\IntOwnN}{9}                    
\newcommand{\IntOwnReduced}{5}              
\newcommand{\IntSlopeRatio}{1.008}          
\newcommand{\IntSlopeLo}{0.85}              
\newcommand{\IntSlopeHi}{1.38}              
\newcommand{\IntSlopeN}{9}                  
\newcommand{\IntSlopeSteeper}{5}            
\newcommand{\GptTwoDropped}{3}              
\newcommand{\GptTwoOldMed}{4}               
\newcommand{\GptTwoDiff}{12}                